%% file: iclr2015.tex
\documentclass{article} 
\usepackage{iclr2015,times}
\usepackage{hyperref}
\usepackage{url}
\usepackage{natbib}
\usepackage{amsmath} 
\usepackage{hyperref}
\usepackage{gensymb} 
\usepackage{caption,subcaption}
\usepackage[pdftex]{graphicx}

\title{Denoising autoencoder with modulated\\
  lateral connections learns invariant\\
  representations of natural images}

\author{
Antti Rasmus and Tapani Raiko\\
Aalto University\\
Finland \\
\texttt{\{antti.rasmus,tapani.raiko\}@aalto.fi} \\
\And
Harri Valpola \\
ZenRobotics Ltd. \\
Vilhonkatu 5 A, 00100 Helsinki, Finland \\
\texttt{harri@zenrobotics.com} \\
}

\newcommand{\h}{\mathbf{h}}

\iclrfinalcopy 

\begin{document}

\maketitle

\begin{abstract}
Suitable lateral connections between encoder and decoder are shown to
allow higher layers of a denoising autoencoder (dAE) to focus on
invariant representations. In regular autoencoders,
detailed information needs to be carried through the highest
layers but lateral connections from encoder to decoder relieve this pressure.
It is shown that abstract invariant features can be translated to detailed
reconstructions when invariant features are allowed to modulate the strength
of the lateral connection.
Three dAE structures with modulated and additive lateral connections,
and without lateral connections
were compared in experiments using real-world images.
The experiments verify that adding modulated lateral connections to the model
1) improves the accuracy of
the probability model for inputs, as measured by denoising
performance; 2) results in
representations whose degree of invariance grows faster towards the
higher layers;
and 3) supports the formation of diverse invariant poolings.
\end{abstract}

\section{Introduction}

Denoising autoencoders \citep[dAE;][]{Vincent-TR1316}
provide an easily accessible 
method for unsupervised learning of representations since training can be based 
on simple back-propagation and a quadratic error function.
An autoencoder is built from two mappings:
an encoder that maps corrupted input data to features, and a decoder that 
maps the features back to denoised data as output. Thus, in its basic form, 
autoencoders need to store all the details about the input in its 
representation.

Deep learning has used unsupervised pretraining
\citep[see][for a review]{bengio2013review}
but recently purely supervised learning has become the dominant approach
at least in cases where a large number of labeled data is available
\citep[e.g.,][]{ciresan:2010,Krizhevsky2012NIPS}.

One difficulty with combining autoencoders with supervised learning is that
autoencoders try to retain all the information whereas supervised learning
typically loses some. For instance,
in classification of images, spatial pooling of activations throws away some of 
the location details while retaining identity details. In that sense, 
the unsupervised and supervised training are pulling the model in very 
different directions. 

From a theoretical perspective, it is clear that unsupervised learning
must be helpful at least in a semi-supervised setting. Indeed,
\citet{Kingma2014NIPS} obtained very promising results with
variational autoencoders. This raises hopes that the same could be
achieved with simpler dAEs.

\begin{figure}[tb] \begin{center}
\begin{subfigure}[b]{0.31\textwidth}
	\includegraphics[width=\linewidth]{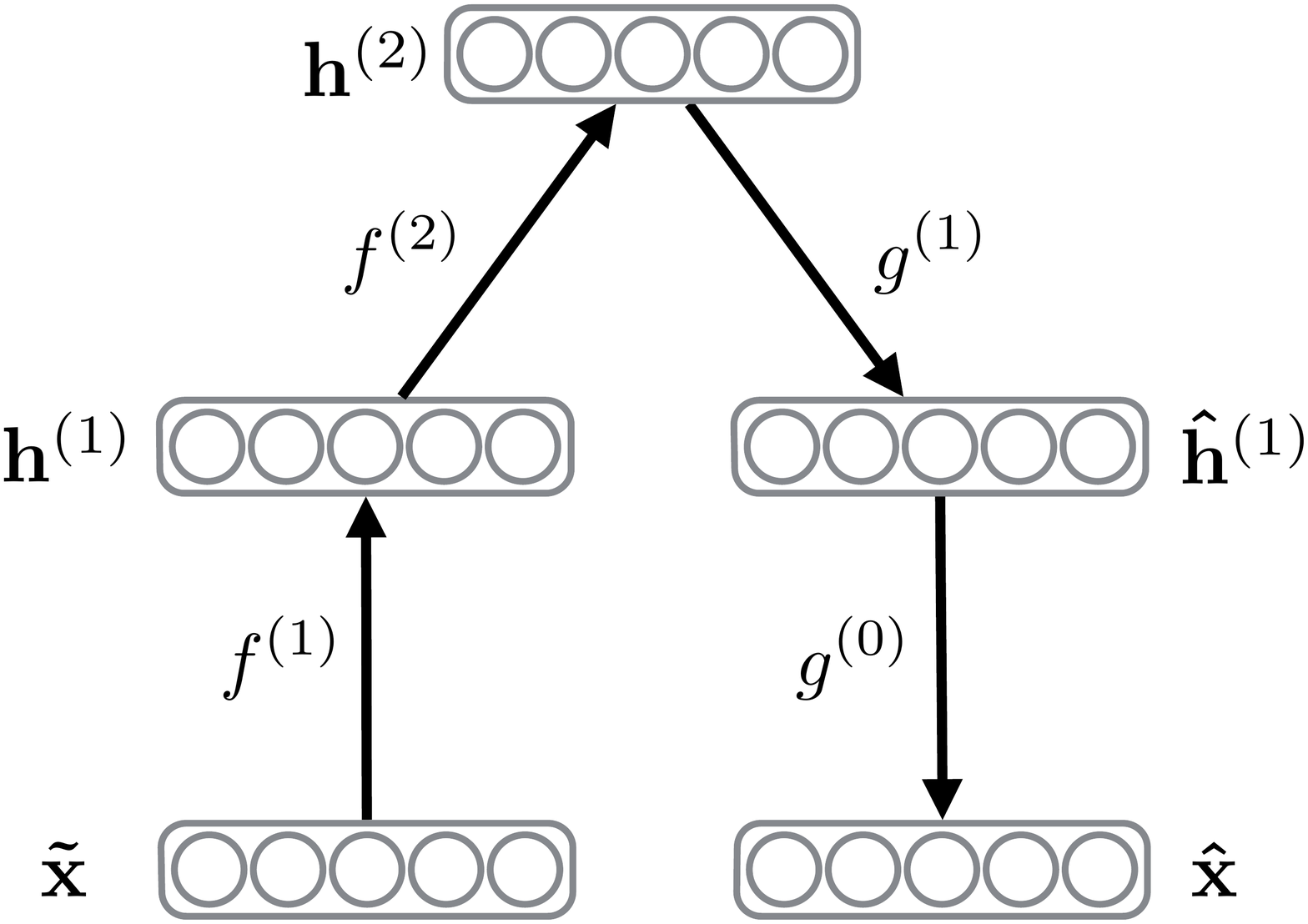}
\caption{Basic autoencoder}\label{arch_autoencoder}
\end{subfigure}
\begin{subfigure}[b]{0.31\textwidth}
	\includegraphics[width=\linewidth]{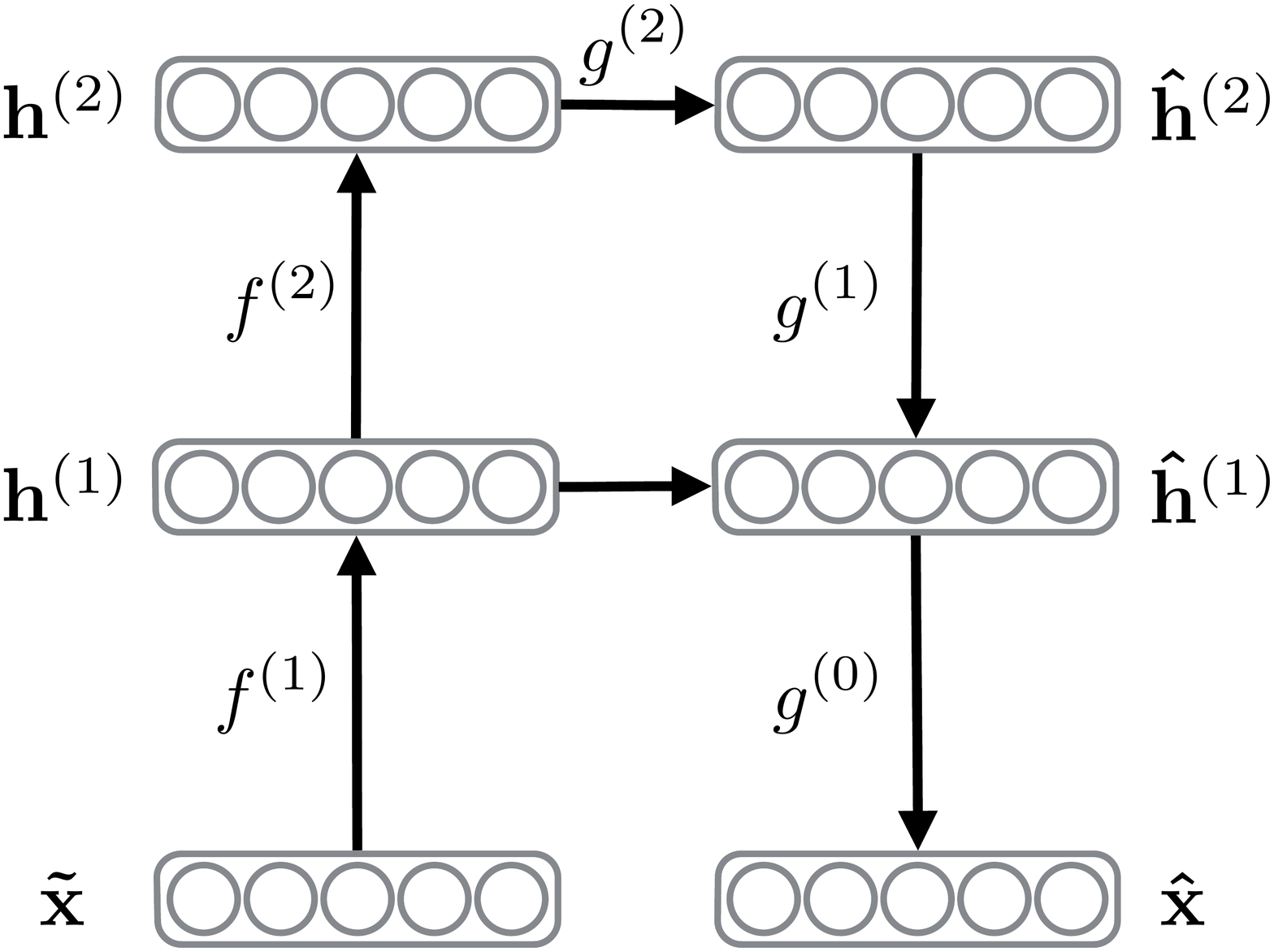}
\caption{Lateral connections}\label{arch_ladder}\end{subfigure}
\begin{subfigure}[b]{0.31\textwidth}
	\includegraphics[width=\linewidth]{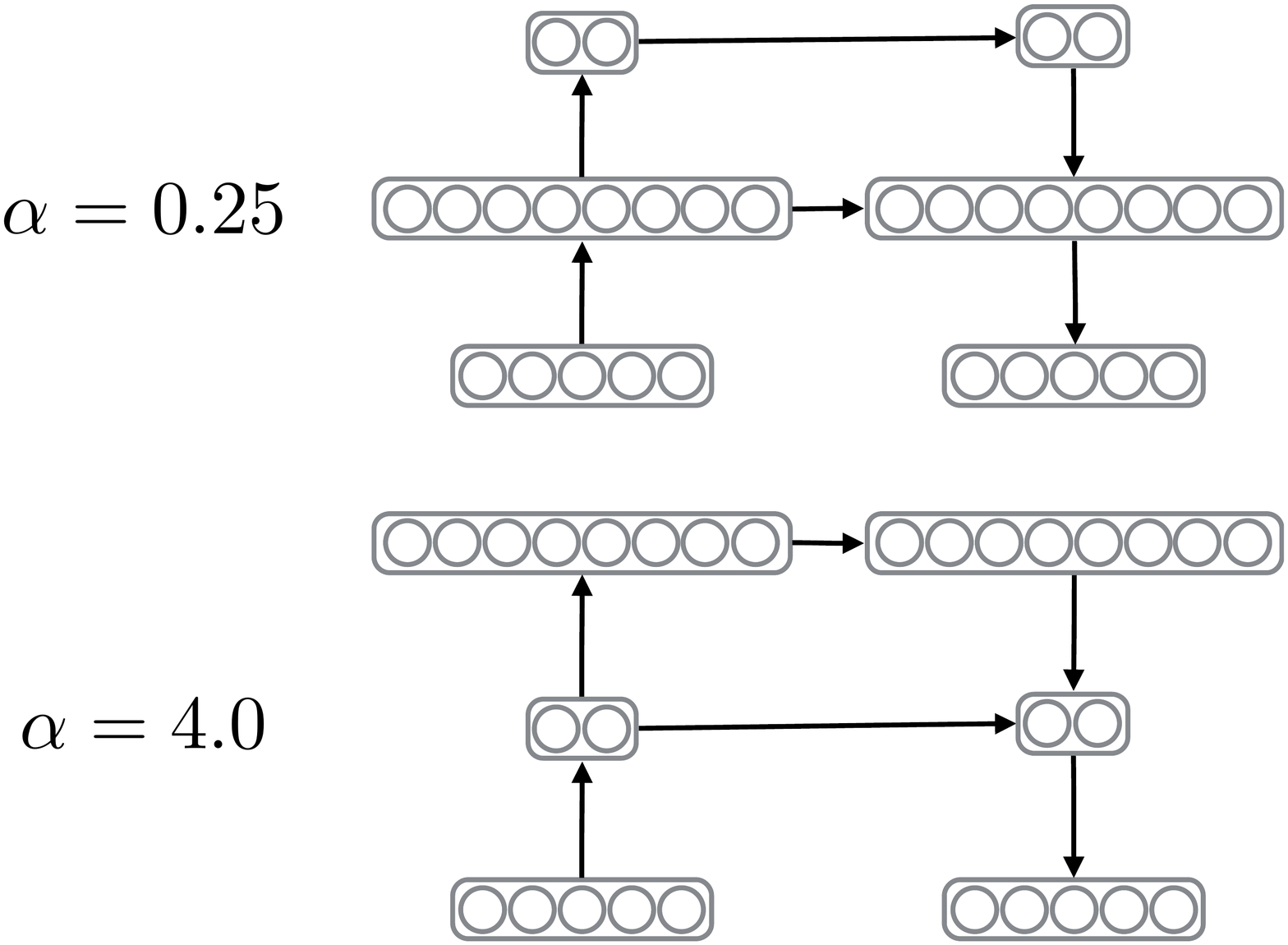}
\caption{Layer size ratio}\label{arch_ratio}\end{subfigure}
\end{center}
\caption{Examples of two-hidden-layer models: (a) denoising autoencoder and
    (b) Ladder network with lateral connections.
    Illustration of ratio, $\alpha$, of $\h^{(2)}$
    size to $\h^{(1)} $ size (right) with hidden layers of sizes 2 and 8. Ratio
    of $\alpha = 1.0$ corresponds to hidden layers of equal size (a and b).}
\label{architectures}
\end{figure}

Recently, \citet{valpola2015ladder} proposed a variant of 
the denoising autoencoder that can lose information. The novelty is in 
lateral connections that allow higher levels of an 
autoencoder to focus on invariant abstract features
and in layer-wise cost function terms that allow the network to learn
deep hierarchies efficiently.
\citet{valpola2015ladder} hypothesized that modulated lateral connections
support the development of invariant features and provided initial results
with artificial data to back up the idea.
As seen in Figure~\ref{architectures},
information can flow from the input to the output 
through alternative routes, and the details no longer need to be stored 
in the abstract representation. This is a step closer to being compatible with supervised
learning which can select which types of invariances and abstractions are 
relevant for the task at hand.

In this paper, we focus on investigating the effects of lateral connections.
We extend earlier results with experiments using natural image data
and make
comparisons with regular denoising autoencoders without lateral
connections in Section~\ref{sec:experiments}. We show the following:
\begin{itemize}
\item The proposed structure attains a better model of the data as
  measured by ability to denoise. There are good reasons to believe that
  this indicates that the network has captured a more accurate
  probabilistic model of the data since denoising is one way of
  representing distributions \citep{bengio2013generalized}.
\item Including the modulated lateral connections changes the optimal  
  balance between the sizes of layers from balanced to bottom heavy.
\item The degree of invariance of the representations grows towards
  the higher levels in all tested models but
  much faster with modulated lateral connections.
  In particular, the higher levels
  of the model seem to focus entirely on invariant
  representations whereas the higher levels of the regular autoencoder
  have a few invariant features mixed with a large number of details.
\item Modulated lateral connections guide the layer above them
  to learn various types of poolings. The pooled neurons participate
  in several qualitatively different poolings that are each selective and
  invariant to different aspects of the input.
\end{itemize}

\subsection{Invariant features through pooling}
\label{sec:invariance-bg}

There are typically many sources of
variation which are irrelevant for classification. For example, in
object recognition from images, these sources could include position,
orientation and scale of the recognized object and illumination
conditions. In order to correctly classify new samples, these sources
of variation are disregarded while retaining information
needed for discriminating between different classes. In other
words, classification needs to be invariant to the irrelevant
transformations.

A simple but naive way to achieve such invariance would be to list all the
possible realizations of objects under various transformations, but
this is not usually practical due to the vast amounts of possible
realizations. Also, this does not offer any generalization to new
objects.

A useful solution is to split the generation of
invariance into more manageable parts by representing the inputs in
terms of features which are each invariant to some types of
transformations. Since different kinds of objects can be represented
by the same set of features, this approach makes it possible to generalize
the invariances to new, unseen objects.

So rather than listing all the possible realizations of individual
objects, we can list possible realizations of their constituent
features. Invariance is achieved by retaining only the information
about whether any of the possible realizations is present and
discarding information about which one exactly. Such an operation is known as
pooling and there are various ways of implementing it. In
the case of binary inputs, OR operation is a natural choice. There are
many ways to generalize this to continuous variables, including
maximum operation \citep{Riesenhuber:99} or summation
followed by a concave function \citep{fukushima-neocognitron}.

While pooling achieves invariance to irrelevant transformations, the
features also need to be discriminative along relevant dimensions.
This selectivity is often achieved by coincidence detection, i.e.
by detecting the simultaneous presence of multiple input features. In the
binary case, it can be implemented by an AND operation.
In the continuous case, possibilities include extensions of the
AND operation, such as product or summation followed by a convex
function, but also lateral inhibition (i.e., competition) among a set
of feature detectors. All of these operations typically produce sparse
coding where the output features are sensitive to specific
combinations of input features. Since this type of coding tends to
increase the number of features, it is also known as feature
expansion.

The idea of alternating pooling and feature expansion dates at least back to
\citet{hubel1962receptive} who found that the early stages of visual
processing in the cat cerebral cortex have alternating steps of
feature expansion, implemented by lateral competition among so called
simple cells, and invariance-generating poolings by so called complex
cells. In such hierarchies of alternating steps, the degree of invariance
grows towards the higher levels.
Cortical processing also includes various normalizations,
a feature which has also been included in some models
\citep[e.g.,][]{fukushima-neocognitron}.

There are various ways of finding poolings that generate invariances
(from specialized to general):
\begin{enumerate}
\item Invariance by design. For instance, invariance to translation
  and small deformations is
  achieved by pooling shifted versions of a feature
  \citep{fukushima-neocognitron}. Similar pooling operations are now popular
  in convolutional neural networks \citep[see, e.g.,][]{Schmidhuber-review}.
\item Invariance to hand-crafted transformations. The transformations
  are applied to input samples (e.g., image patches can be shifted,
  rotated, scaled or skewed, and colors or contrast can be modified) and
  pooling is then learned by requiring the output to stay constant over the
  transformation. This category includes supervised learning from inputs
  deformed by various transformations.
\item Invariance to transformations over time.
  Relies on nature to provide transformations as sequences
  during which the underlying feature (e.g., identity of object)
  changes slower than the transformation \citep[e.g.,][]{Foldiak:1991}.
\item Invariance by exploiting higher-order correlations within
  individual samples.
  This is how supervised learning can find poolings:
  target labels correlate nonlinearly with inputs.
  There are also unsupervised methods that 
  can do the same.  For example, subspace ICA can find
  complex-cell like poolings for natural images \citep{hyvarinen2000emergence}.
\end{enumerate}
We focus on the last type: exploiting higher-order correlations.
Very few assumptions are made so the method is very general
but it is also possible to combine this approach with supervised learning
or any of the more specialized ways of generating invariances.

\subsection{Denoising autoencoders}

Autoencoder networks have a natural propensity to conserve
information and are therefore well suited for feature expansion.
Autoencoder networks consist of two parameterized functions, encoding $f$ and
decoding $g$.  Function $f$ maps from input space $\mathbf x$ to feature space
$\h$, and $g$ in
turn maps back to input space
producing a reconstruction, $\mathbf{\hat x}$, of the original
input, when the training criterion is to minimize the reconstruction error. This
enables learning of features in an unsupervised manner.

Denoising autoencoder \citep{Vincent-TR1316} is a variant of the traditional
autoencoder, where the input $\mathbf x$ is corrupted with noise
and the objective of the network is to reconstruct the
original uncorrupted input $\mathbf{x}$ from the corrupted
$\mathbf{\tilde x}$.
\citet{bengio2013generalized} show that denoising autoencoders implicitly 
estimate the data distribution as the asymptotic distribution of 
the Markov chain that alternates between corruption
and denoising. This interpretation provides a solid probabilistic foundation 
for them.
Consequently, the denoising teaching criterion enables learning of over-complete
representations, a property which is crucial for adding lateral connections
to an autoencoder.

As with normal feedforward networks, there are various options for choosing
the cost function but, in the case of continuous input variables, a simple choice
is
\begin{align}
  \mathcal{C} = \|\mathbf{\hat x} - \mathbf x\|^2
  = \|g(f(\mathbf{\tilde x})) - \mathbf x\|^2
  \, . \label{eq:cost}
\end{align}

\newcommand{\La}{^{(L)}} 
\newcommand{\lap}{^{(l+1)}} 
\newcommand{\xt}{\mathbf{\tilde x}}
\newcommand{\hh}{\mathbf{\hat h}}
\newcommand{\bv}{\mathbf{b}}
\newcommand{\W}{\mathbf{W}}
\newcommand{\la}{^{(l)}} 
\newcommand{\var}{\mathrm{var}}

Denoising autoencoders can be used to build deep architectures either by
stacking several on top of each other and training them in a greedy layer-wise
manner \citep[e.g.,][]{Bengio-nips-2006} or by chaining several encoding and
decoding functions and training all layers simultaneously. For $L$ layers and
encoding functions $f\la$, the encoding path would compose as
$f = f^{(L)} \circ f^{(L-1)} \circ \dots f^{(1)}$.

We denote the intermediate feature vectors by $\h\la$ and corresponding decoded,
denoised, vectors by $\hh \la$. Figure~\ref{arch_autoencoder} depicts such a
structure for $L=2$. Encoding functions are of the form
\begin{equation}
\newcommand{\hver}{\h^{(l-1)}} 
\mathbf{h}^{(l)} = f^{(l)}(\hver) = \phi\big(\W\la_f \hver + 
\bv_f^{(l)}\big),\quad 1 \leq l \leq L,
\label{eq:f}
\end{equation}
starting from $\h^{(0)}=\mathbf{\tilde x}$.
The corresponding decoding functions are
\newcommand{\hver}{\hh\lap} 
\begin{align}
    \hh\La &= \h\La \label{eq:dae_gL} \\
    \hh\la &= g\la(\hver) = \phi \big(\W_g\la \hver + 
\bv_g\la\big),\quad 1 \leq l \leq L-1,
\label{eq:dae_g} \\
    \mathbf{\hat x} &= g^{(0)}(\hh^{(1)}) = \W_g^{(0)} \hh^{(1)} + \bv_g^{(0)}.
\label{eq:dae_g0}
\end{align}

Function $\phi$ is the activation function and typically left out from the lowest
layer.

\section{Experiments}
\label{sec:experiments}

The tendency of regular autoencoders to preserve information seems to
be at odds with the development of invariant features which relies on
poolings that selectively discard some types of information. Our goal
here is to investigate the hypothesis that suitable lateral
connections allow autoencoders to discard details from higher layers
and only retain abstract invariant features because the decoding functions
$g\la$ can recover the discarded details from the encoder.

We compare three different denoising autoencoder
structures:
basic denoising autoencoder and two variants with lateral
connections.
We experimented with various model definitions prior to deciding the ones
defined in Section \ref{sec:lateral} because there are multiple ways to
merge lateral connections.

\subsection{Models with Lateral Connections}
\label{sec:lateral}

We add lateral connections from $\h\la$ to $\hh\la$ as seen in 
Figure \ref{arch_ladder}. Autoencoders trained without noise would 
short-circuit the input and output with an identity mapping.
Now that input contains noise, there is pressure to find
meaningful higher-level representations that capture regularities
and allow for denoising. 
Note that the encoding function $f$ has the same form as before in
Eq.~\eqref{eq:f}.

\subsubsection{Additive lateral connections}

As the first version, we replace the decoding functions in
Eq.~(\ref{eq:dae_gL}--\ref{eq:dae_g}) with
\begin{align}
\hh\La &=  g\La (\h\La) = (\h\La + \bv_a\La) \odot \sigma 
\big(\mathbf a\La 
\odot \h\La + \bv_b\La \big) \, , \label{eq:add_gL} \\
 \hh\la &= g\la(\h\la, \hver) = (\h\la + \bv_a\la) \odot 
\sigma 
\big(\mathbf a \la
\odot \h\la + \bv_b\la \big) + \phi\big(\W_g\la\hver + 
\bv_g\la\big) \, , \label{eq:add_gl}
\end{align}
where $\odot$ is the element-wise (i.e., Hadamard)
product, $\mathbf a\la$, $\bv_a\la$, 
and $\bv_b\la$ are learnable parameter vectors along with weights and biases,
and $\sigma$ is a sigmoid function to ensure that the
modulation stays within reasonable bounds.
Function $g^{(0)}$ stays affine as in Eq.~\eqref{eq:dae_g0}.
The functional form of Eq.~\eqref{eq:add_gL}, with element-wise decoding,
is motivated by element-wise denoising functions that are used in
denoising source separation \citep{Sarela:2005} and corresponds to
assuming the elements of $\h$ independent a priori.

\subsubsection{Modulated lateral connections}

Our hypothesis is that an autoencoder can learn invariances
efficiently only if its decoder can make good use of them.
\citet{valpola2015ladder} proposed connecting the top-down
mapping inside the sigmoid term of Eq.~\eqref{eq:add_gl}, a choice
motivated by optimal denoising in hierarchical variance models.

Our final proposed model includes the encoding functions in Eq.\ \eqref{eq:f}, 
the top connection $g^{(L)}$ in Eq.\ \eqref{eq:add_gL}, bottom 
decoding function $g^{(0)}$ as in Eq.~\eqref{eq:dae_g0}, but the middle decoding 
functions are defined as
\begin{align}
g\la(\h\la, \hver) &= (\h\la + \bv_a\la) \odot \sigma \big(\mathbf a
\la \odot \h\la + \W_g\la \hver + \bv_b\la\big), \quad 1 \leq l \leq L 
- 1.
\label{eq:mod_g}
\end{align}
In contrast to additive lateral connection in Eq.~\eqref{eq:add_gl},
the signal from the abstract layer $\hver$ is used to modulate the lateral
connection so that the
top-down connection
has moved from $\phi(\cdot)$ to $\sigma(\cdot)$,
and the bias $\bv_g\la$ has dropped because it is redundant
with the bias term $\bv_b\la$ in $\sigma(\cdot)$.

Modulated (also known as gated or three-way)
connections have been used in autoencoders 
before but in a rather different context. \citet{memisevic_gradient_based_2011} 
uses a weight tensor to connect two inputs to a feature. We connect the 
$i$th component of $\h\la$ only to the $i$th component of $\hh\la$, keeping
the number of additional parameters small.

\subsection{Training procedure}

In order to compare the models, we optimized each model structure
constraining to 1 million parameters and 1 million mini-batch updates
to find the best denoising performance, that is, the lowest reconstruction
cost $\mathcal{C}$ in Eq.~\eqref{eq:cost}.
The better the denoising performs the better the
implicit probabilistic model is. The tasks is the same for all models, so the
comparison is fair.
With two-layer models ($L=2$) the focus was to find the optimal size for
layers, especially the ratio of $\h^{(2)}$ size to $\h^{(1)}$ size
(see Fig.~\ref{arch_ratio} for the definition of the ratio $\alpha$).
All the models use rectified linear unit as the activation
function $\phi(\cdot)$ and the noise was Gaussian with zero mean and
standard deviation scaled to be $50\%$ of that of the data.

In order to find the best possible baseline for comparison,
we evaluated weight tying for autoencoder without lateral connections
, that is, $\W_g\la = {\W_f\lap}^T$,
and noticed that tying weights resulted in faster convergence and thus better
denoising performance than without weight tying. However, when combining both
so that weights are tied in the beginning and untied for the latter half of
the training, denoising performance improved slightly (by $1\%$), but did not affect
the relative difference of various models.
Since weight tying had only negligible impact on the results, but it complicates
training and reproducibility,
we designed the experiments so that all models use tied weights
counting tied weights as separate parameters.
Results of the denoising performance are described in
Section~\ref{sec:denoising_performance}.

We performed the analysis on natural images because the
invariances and learned features are easy to visualize, we know
beforehand that such invariances do exist, and because computer
vision is an important application field in its own right.
We used $16 \times 16$ patches from two image datasets: CIFAR-10
\citep{KrizhevskyHinton2009} and natural images used by
\citet{Olshausen+Field-1996}
\footnote{Available at \url{https://redwood.berkeley.edu/bruno/sparsenet/}}.
We refer to this dataset as O\&F.

The training length was limited to 1 million mini-batch updates
with mini-batch of size 50 and learning rate was adapted with ADADELTA
\citep{zeiler2012adadelta}.
The best variants of each model were then trained
longer, for 4 million updates, and further analyzed to determine invariance of
the learned representations. This is described and reported in
Section~\ref{sec:invariance}.
Supplementary material provides more details about data preprocessing,
division between training and validation sets, training procedure,
hyperparameters used with ADADELTA, and how weights were initialized.

We also tried stacked layer-wise training by training each layer for 500,000
updates followed by fine-tuning phase of another 500,000 updates such that the
total number of updates for each parameter equals to 1 million. We also tried
a local cost function and noise source on each layer or using the global
cost, but we did not find any such stacked training beneficial compared to the
direct simultaneous training of all layers, which is what we report in this paper.

\subsection{Denoising performance}
\label{sec:denoising_performance}

\begin{figure}[tb]
\begin{center}
\includegraphics[width=0.49\linewidth]{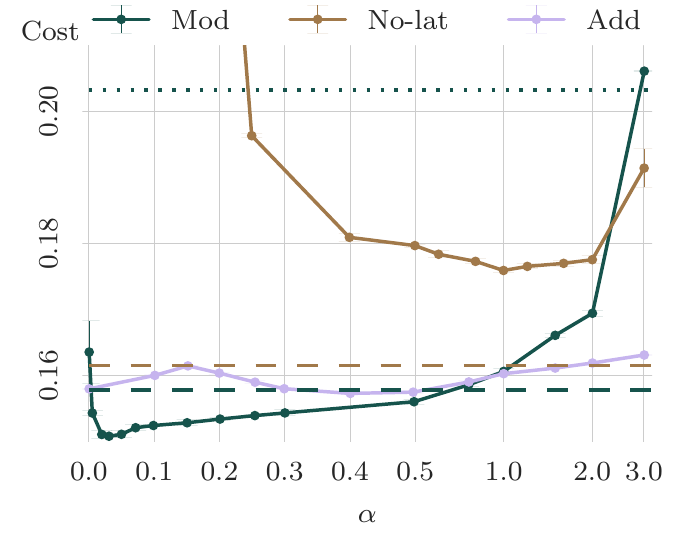}
\includegraphics[width=0.49\linewidth]{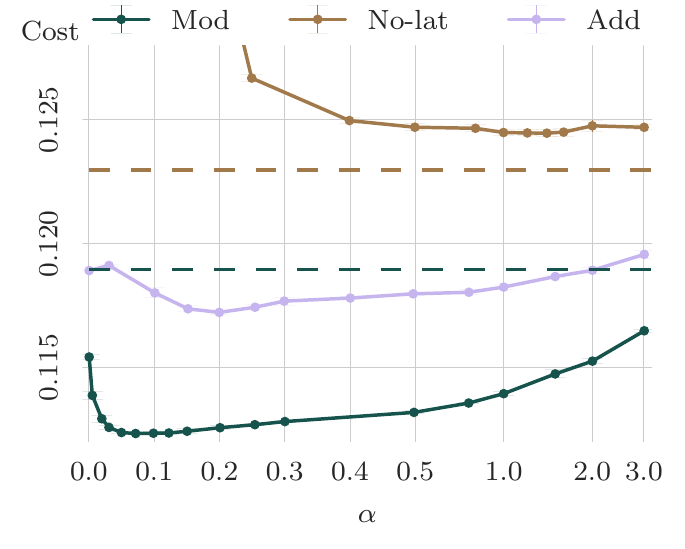}
\end{center}
\caption{The best validation cost per element as a function of $\alpha$,
the ratio of $\h^{(2)}$ size
to $\h^{(1)}$ size, for CIFAR-10 (left) and O\&F (right) datasets.
Dotted line is the result of linear denoising
(model Linear in Table~\ref{results}),
two dashed lines represent the
denoising performance of one-layer models $(L=1)$ with and without lateral connections
according to their
colors. Note that Add and Mod models are identical when $L=1$.
The scale of the horizontal axis is linear until 0.5 and logarithmic
after that.}
\label{ratio}
\end{figure}

The results of denoising performance for models with one and two layers is
presented in the Figure~\ref{ratio} which shows the lowest reconstruction cost
on a validation dataset for both datasets.
Each configuration was trained 5 times with different random initialization for
confidence bounds calculated as corrected sample standard deviation of the
lowest reconstruction cost.

The best
performing No-lat model (autoencoder without lateral connections)
is a one-layer model and is shown in dashed line. The best
two-layer No-lat models
in CIFAR-10 and O\&F have ratios of $\alpha_{min}=1.0$ and
$\alpha_{min}=1.4$, respectively.
Since No-lat autoencoders need to push all the information through each
layer, it is intuitive that very narrow bottlenecks are bad for the model,
that is, large and small ratios perform poorly. A further study of why the 
optimal ratio
is larger for O\&F revealed that the lower layer can be smaller because the
effective dimensionality in terms of principal components is lower for O\&F
compared to CIFAR-10.
Second layer is not beneficial with No-lat model given
the parameter number constraint.

Mod (modulated lateral connections) model benefits from the second layer and
works best when the ratio is small, namely $\alpha_{min}=0.03$ and
$\alpha_{min}=0.12$ for CIFAR-10 and O\&F,
respectively. The second layer does not hurt or benefit Add 
(additive lateral connection) model
significantly
and its performance is between No-lat and Mod models.
The results are also presented as numbers in Table~\ref{results}
in supplementary material.

\begin{figure}[tb]
\begin{center}
\begin{subfigure}[b]{0.30\textwidth}
	\includegraphics[width=\textwidth]{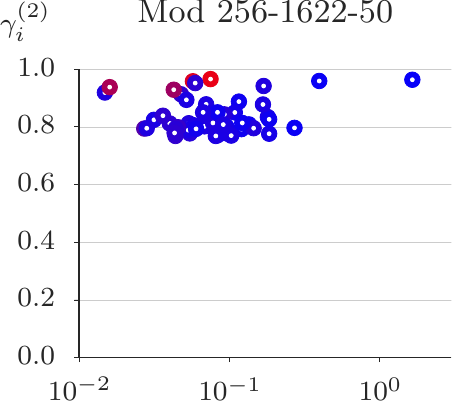}
\end{subfigure}
\begin{subfigure}[b]{0.30\textwidth}
	\includegraphics[width=\textwidth]{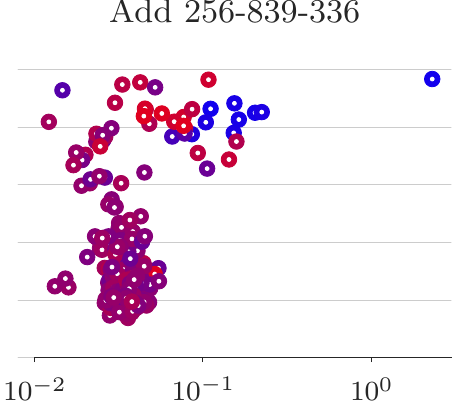}
\end{subfigure}
\begin{subfigure}[b]{0.30\textwidth}
	\includegraphics[width=\textwidth]{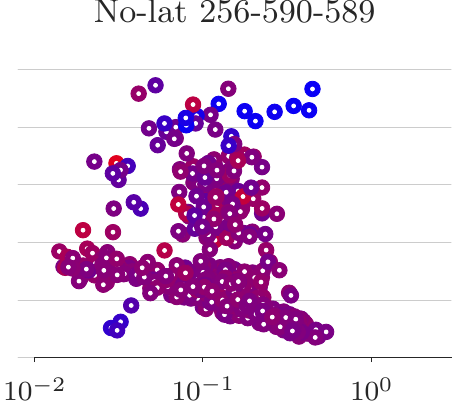}
\end{subfigure}
\end{center}
\caption{Translation invariance measure of neurons on $h^{(2)}$ as a function of
significance. Color indicates the average sign of the connected weights from negative
(blue) to positive (red).
Best viewed in color. See Section~\ref{sec:invariance} for more details.}
\label{invariance}
\end{figure}

\subsection{Resulting invariant features}
\label{sec:invariance}

Practically no prior information about poolings was incorporated in
either the model structure or treatment of training data. This means
that any invariances learned by the model must have been present in
the higher-order correlations of the inputs, as explained in
Section~\ref{sec:invariance-bg}. It is well known that such
invariances can be developed from natural images
\citep[e.g.,][]{hyvarinen2000emergence} but the question is, how well
are the different model structures able to represent and learn these
features.

To test this, we generated sets of rotated, scaled and translated images
and measured how invariant the activations $\h\la$ are for
each type of transformation separately.
As an example for translation, each set contained 16 translated images
(from a $4 \times 4$ grid).
For each set $s$ in a given transformation type, we calculated the mean activation
$\langle\h\la\rangle_s$ and compared their variances with the
variance $\var\{\h\la\}$
over all samples:
\begin{equation}
  \gamma_i\la = \var\{ \langle h_i\la \rangle_s \} / \var\{h_i\la\} \, .
  \label{eq:gamma}
\end{equation}

From the definition it follows that $0 \leq \gamma_i\la \leq 1$. If
the feature is completely invariant with respect to the transformation,
$\gamma_i\la$ equals one\footnote{Various measures of invariance have
  been proposed \citep[e.g.,][]{goodfellow2009measuring}. Our
  invariance measure $\gamma$ is closely related to autocorrelation
  which has been used to measure invariance, e.g., by
  \citet{Dosovitskiy-invariance}. The set of translated patches
  includes pairs with various translations and therefore our measure
  is a weighted average of the autocorrelation function. The wider the
  autocorrelation, the larger the measure $\gamma$.}.
  As the overall conclusions are similar to all tested transformations,
  the main text concentrates on the translation invariance.
Results for other invariances are reported in supplementary material,
Section~\ref{sec:other_invariances}.
The average layer-wise invariance, $\gamma\la = \langle \gamma_i\la \rangle_i$,
grows towards the higher layers in all models but much
faster in the best Mod models than in others\footnote{Our measure
  $\gamma$ can be fooled by copying the same invariant feature to many
  hidden neurons but we verified that this is not happening here: slow
  feature analysis is robust against such redundancy but yields
  qualitatively the same results for the tested networks.},
i.e. for CIFAR-10 the best Mod model has $\gamma^{(0)\dots(2)} = (0.20, 0.31, 0.84)$,
whereas for the best No-lat $\gamma^{(0)\dots(2)} = (0.20, 0.23, 0.30)$.
All $\gamma\la$ values
are reported in Table~\ref{results} in supplementary material.

To further illustrate this, we plotted
in Figure~\ref{invariance} the
invariance measures $\gamma_i^{(2)}$ for the best variant of each model.
In each plot, dots correspond to hidden
neurons $h_i^{(2)}$ and the color reflects the average sign of the
encoder connection to that neuron (blue for negative, red for positive).
The horizontal axis is the significance of a neuron, a measure of how much
the model uses that hidden neuron, which is defined and analyzed in
supplementary material, Section~\ref{sec:mappings}.

There are two notable observations. First, all $h_i^{(2)}$ neurons of
Mod model are highly invariant, whereas the other models have only few
invariant neurons and the vast majority of neurons have very low invariance.
For No-lat model, invariance seems to be even smaller for those neurons that
the model uses more. Moreover, we tested that the second layer of Mod model
stays highly invariant even if the layer size is increased
(shown in Figure~\ref{invariance_size}
in supplementary material).
Second, invariant neurons tend to have far more and stronger negative
than positive weights, especially so with Mod and No-lat models.
Since the nonlinearity $\phi(x)$ on
each layer was the rectified linear unit, a convex function which
saturates on the negative side, negative tied weights mean that the
network flipped these functions into $-\phi(-x)$, that is, concave
functions that saturate on the positive side resembling OR operation.
This interpretation is further discussed in supplementary material,
Section~\ref{sec:negative_weights}.

\subsection{Learned poolings}

\begin{figure}[tb]
\begin{center}
\begin{subfigure}[b]{1.0\textwidth}
    \begin{minipage}[c]{0.1\textwidth}
    \caption{}
    \end{minipage}\hfill
    \begin{minipage}[c]{0.9\textwidth}
    \includegraphics[width=\textwidth]{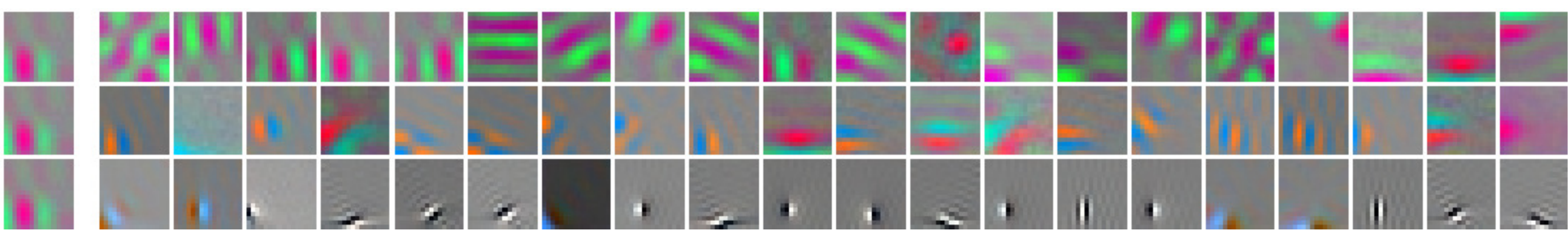}
    \end{minipage}\hfill
\end{subfigure}
\begin{subfigure}[b]{1.0\textwidth}
    \begin{minipage}[c]{0.1\textwidth}
    \caption{}
    \end{minipage}\hfill
    \begin{minipage}[c]{0.9\textwidth}
    \includegraphics[width=\textwidth]{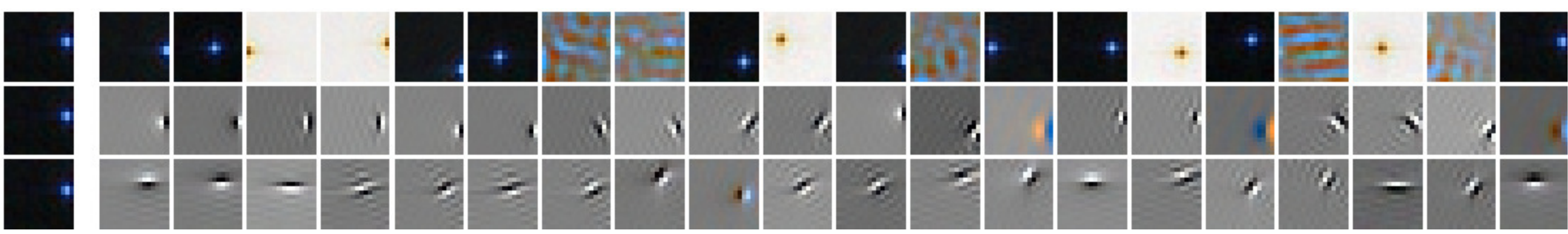}
    \end{minipage}\hfill
\end{subfigure}
\begin{subfigure}[b]{1.0\textwidth}
    \begin{minipage}[c]{0.1\textwidth}
    \caption{}
    \end{minipage}\hfill
    \begin{minipage}[c]{0.9\textwidth}
    \includegraphics[width=\textwidth]{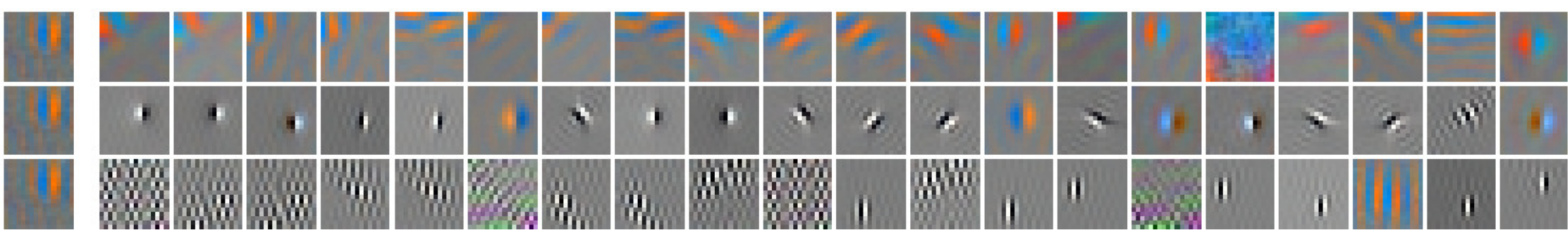}
    \end{minipage}\hfill
\end{subfigure}
\begin{subfigure}[b]{1.0\textwidth}
    \begin{minipage}[c]{0.1\textwidth}
    \caption{}
    \end{minipage}\hfill
    \begin{minipage}[c]{0.9\textwidth}
    \includegraphics[width=\textwidth]{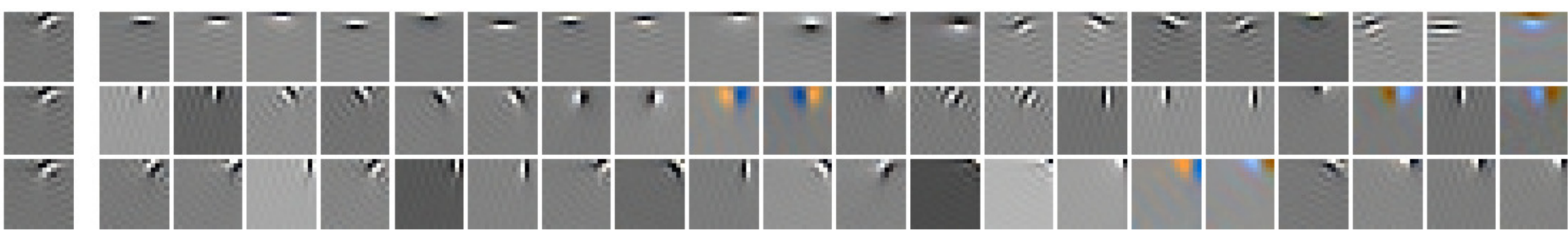}
    \end{minipage}\hfill
\end{subfigure}
\end{center}
\caption{Various pooling groups a neuron. Each set (a)-(d) represents three
    relevant pooling groups the selected neuron $h^{(1)}_a \dots h^{(1)}_d$,
    depicted in the first column, belongs to. Each row represents
    an $h^{(2)}$ pooling neuron by showing the 20 most relevant $h^{(1)}$
    neurons associated with it.
    Poolings were found by selecting a few Layer 1 neurons and following their
    strongest links to Layer 2 to identify the poolings in which they participate.
    Consecutively, the Layer 1 neurons corresponding to each pooling group were
    identified by walking back the strongest links from each Layer 2 neuron.
    Best viewed in color.
    The procedure of choosing features in the plot is also depicted in
	Figure~\ref{pooling_explained} in supplementary material.
}
\label{filters}
\end{figure}

The modulated Mod model used practically all of the second layer neurons
for pooling.
When studying the poolings,
we found that typically every Layer 1 neuron participates in several
qualitatively different poolings.
As can be seen from the
Figure~\ref{filters}, each Layer 1 neuron (shown on the left-most column)
participates in different kinds of poolings each of which is sensitive
to a particular set of features and invariant to other types. For
example, Layer 1 neuron (c) is selective to orientation, frequency and
color but it participates in three different Layer 2 poolings
. The first one is
selective to color but invariant to orientation. The
second one is selective to orientation but invariant to
color. The third one only responds to high frequency and orientation.
More details of the analysis are available in supplementary material.

\section{Discussion}
\label{sec:discussion}

The experiments
showed that invariance in denoising autoencoders increased
towards the higher layers but significantly so only if the decoder had
a suitable structure where details could be combined with invariant features
via multiplicative interactions. Lateral connections from encoder to
decoder allowed the models to discard details from the higher levels but
only if the details and invariant features were combined suitably.

The best models with modulated lateral connections were able to learn a large
number of
poolings in an unsupervised manner. We tested their invariance
but nothing in the model biased learning in that direction and we
observed invariance and selective discrimination of several different
dimensions such as color, orientation and frequency.

In summary, these findings are fully in line with the earlier
proposition that the unsupervised denoising autoencoder with modulated
lateral connections can work in
tandem with supervised learning because, as we have shown here for the
first time, the higher layers of the model have the
ability to focus on abstract representations and, unlike regular
autoencoders, should therefore be able to discard details if
supervised learning deems them irrelevant.
It now becomes possible to combine autoencoders with the popular
supervised learning pipelines which include in-built pooling
operations \citep[see, e.g.,][]{Schmidhuber-review}.

There are multiple ways to extend the work, including 1)
explicit bias towards invariances; 2) sparse structure such as
convolutional networks, making much larger scale models and deeper
hierarchies feasible; 3) dynamic models; and 4) semi-supervised
learning.



\bibliography{iclr2015,lisa-strings,lisa-nn}
\bibliographystyle{iclr2015}

\newpage
\section{Supplementary material}

\begin{table}[bt]
\caption{Denoising performance and translation invariance measure of selected models.
All
models have exactly the same input layer and data so the average invariance is the
same for all models, e.g. for CIFAR-10, $\gamma^{(0)}=0.20$.
}
\label{results}
\begin{center}
\begin{tabular}{lllllllll}
\input{results.tex}
\end{tabular}
\end{center}
\end{table}

\subsection{Preprocessing of data}

Patches of size $16 \times 16$ were sampled randomly and continuously during
training. Separate test images were put aside for testing generalization
performance: the last 10,000 samples for CIFAR-10 and the sixth image of O\&F
dataset. Continuous sampling allows generation of millions of data samples
alleviating overfitting problems. O\&F dataset was already (partially)
whitened so no
further preprocessing was applied to it. RGB color patches of CIFAR-10 were
whitened and dimensionality was reduced to 256 with PCA to match the
dimensionality of grayscale images of O\&F. Despite dimensionality reduction,
$99\%$ of the variance was retained.

\subsection{Details of the training procedure}
White additive Gaussian noise of $\sigma_N=0.5$ was used for corrupting the
inputs which were scaled to have standard deviation of $\sigma=1.0$. During
training, ADADELTA was used to adapt the learning
rate with its momentum set to $0.99$, and $\epsilon=10^{-8}$.
All weight vectors were initialized
from normal distribution to have a 
norm of
$1.0$, and orthogonalized. In order to improve the convergence speed of all
models, we centered the hidden unit activations following
\citet{RaikoVL12}: there 
is an auxiliary bias term $\beta$ \citep[Eq.\ (2)]{RaikoVL12}, applied 
immediately after the nonlinearity, that centers the output to zero 
mean.\footnote{We did not use the other transformation $\alpha$ since it would 
have required shortcut connections.}

\subsection{Analysis of mappings}
\label{sec:mappings}
We analyzed the mappings learned by different types of models in
several ways and present here some of the most interesting findings.

First, it turned out that different models had very different
proportions of invariant neurons on $\h^{(2)}$ (invariance measure
$\gamma$ is defined in Eq.~\eqref{eq:gamma}) and we wanted to
understand better what was going on. Some key questions were how
important roles different types of features had and how the invariances
were formed. Second level invariances could be low-frequency features
which are invariant already on the first layer (and thus not
particularly interesting) or formed through pooling Layer 1 neurons.

\begin{figure}[p] \begin{center}
\begin{subfigure}[b]{0.7\textwidth}
	\includegraphics[width=\linewidth]{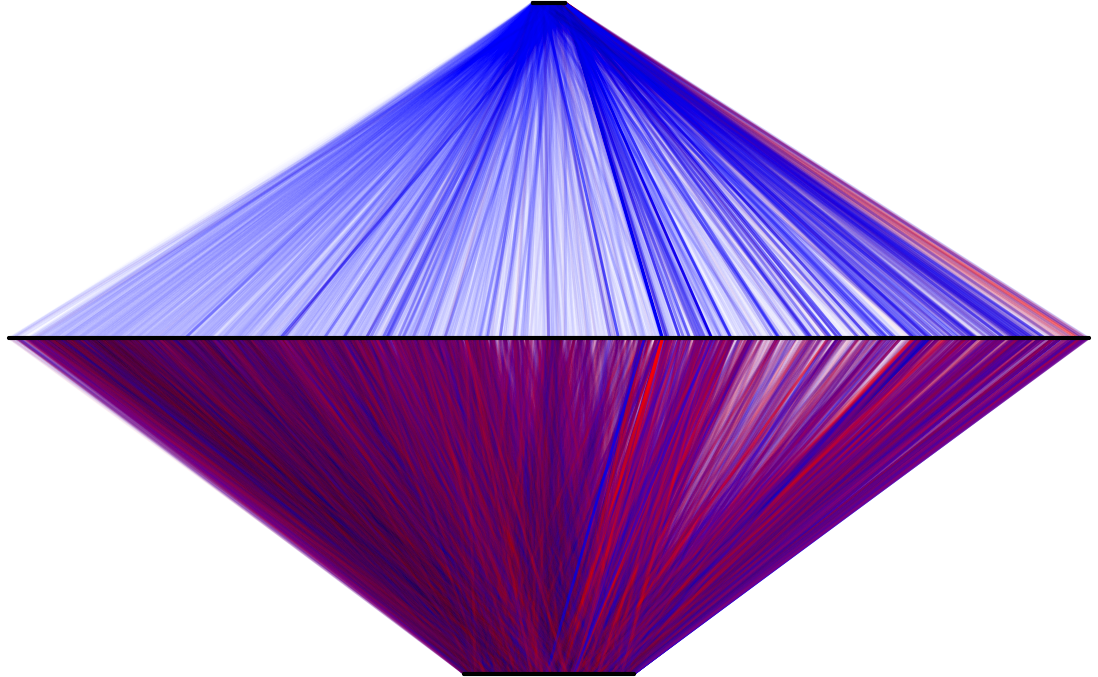}
\caption{Connections of Mod model 256-1622-50}\label{connections_mod}
\vspace{0.5cm}
\end{subfigure}
\begin{subfigure}[b]{0.7\textwidth}
	\includegraphics[width=\linewidth]{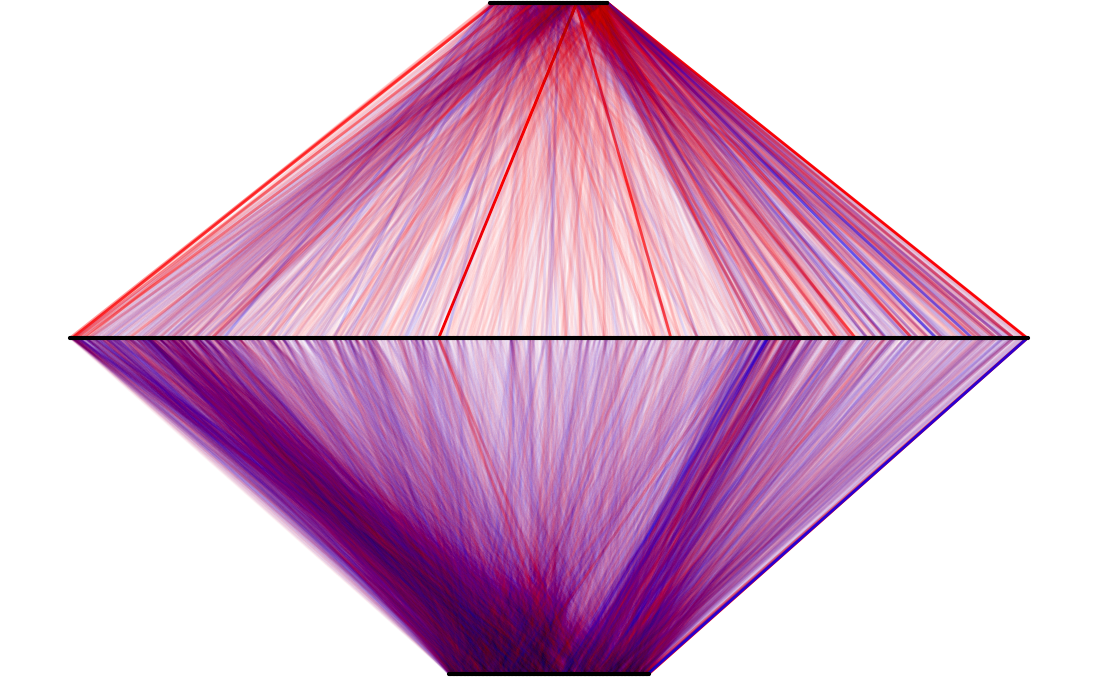}
\caption{Connections of Add model 256-839-336.}\label{connections_add}
\vspace{0.5cm}
\end{subfigure}
\begin{subfigure}[b]{0.7\textwidth}
	\includegraphics[width=\linewidth]{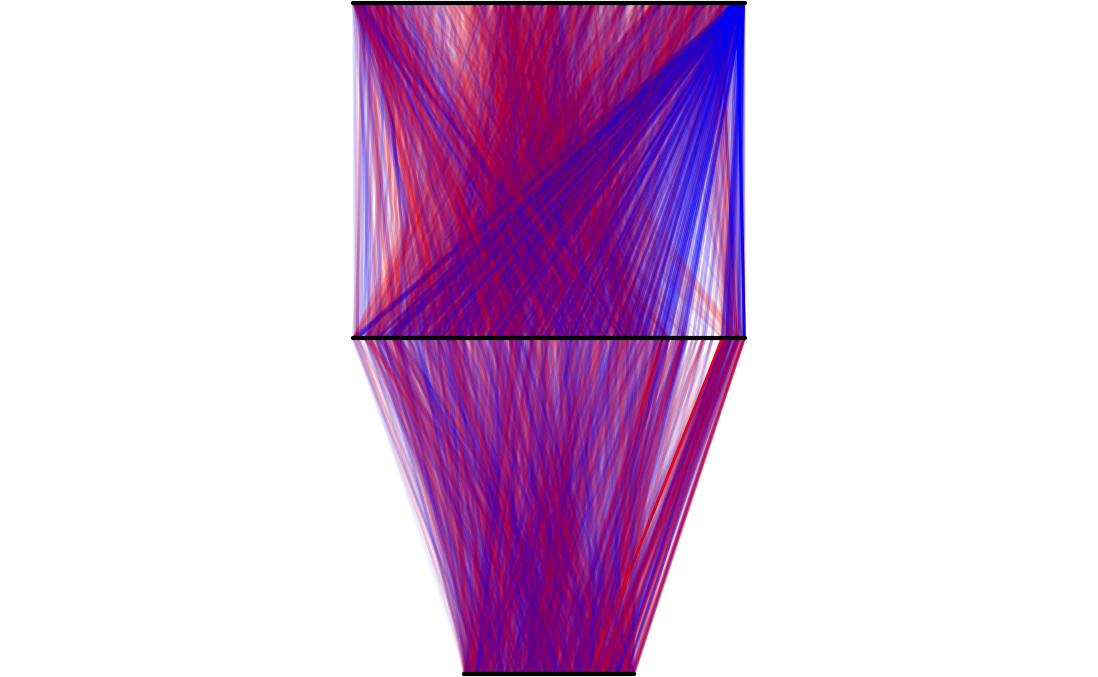}
\caption{Connections of No-lat model 256-590-589.}\label{connections_no589}\end{subfigure}
\end{center}
\caption{Neurons are
      ordered according to increasing translation invariance from left to
      right. Blue denotes negative and red positive weights in
      $\W_g\la$. Strength of the connections depends on the significance
      of the connection (see text for details). Layer 2 neurons of this
      model are also visualized in Figure~\ref{invariance}.
      Best viewed in color.}
\label{connections}
\end{figure}

The first question can be answered by looking at where the connections
are coming from. The connections $\W_g\la$ are visualized in Figures
\ref{connections_mod}--\ref{connections_no589}. The neurons on each
layer have been ordered with respect to
invariance which increases from left to right. The connecting edges
have been colored according to the sign of the connecting weight and
the strength of each edge reflects the significance of the connection.
Significance is defined as the proportion of variance that the
higher-level neuron generates on the lower-level neurons. We initially
tried visualizing simply the magnitudes of the weights but the problem
is that when an input neuron has a low variance or the output neuron is
saturated, a large weight magnitude does not indicate that the
connection is important; it would not make much difference if such a
connection were removed.

\subsubsection{Significance measure}
\label{sec:significance}

When visualizing $\W_g\la$, we first took the squares and scaled them
by the input neuron's variance $\var\{h_i^{(l+1)}\}$. Assuming input
neurons are independent, this quantity reflects the input variance
each neuron $h_j\la$ receives. Depending on the saturation of the
neuron, a smaller or greater proportion of this variance is
transmitted to the actual output variance $\var\{h_j\la\}$. We
therefore scaled all the incoming variances for each $h_j\la$ such
that their sum matches the output variance $\var\{h_j\la\}$. We named
this quantity the significance of the connection and it approximately
measures where the output variance of each layer originates from.
This significance is also depicted in Figure~\ref{invariance} where
the x coordinate is the sum of output significances for each $h_i^{(2)}$.

\subsubsection{Role of negative weights}
\label{sec:negative_weights}
It turned out that the invariant neurons tend to have far more
and stronger negative than positive weights. We visualized this with
color in Figure~\ref{connections}: blue signifies negative and red positive weights. In the
images, the connections are translucent which means that equal number
(and significance) of positive and negative weights results in
purple color.

\begin{figure}[p]
\begin{center}
\begin{subfigure}[b]{0.49\textwidth}
	\includegraphics[width=\linewidth]{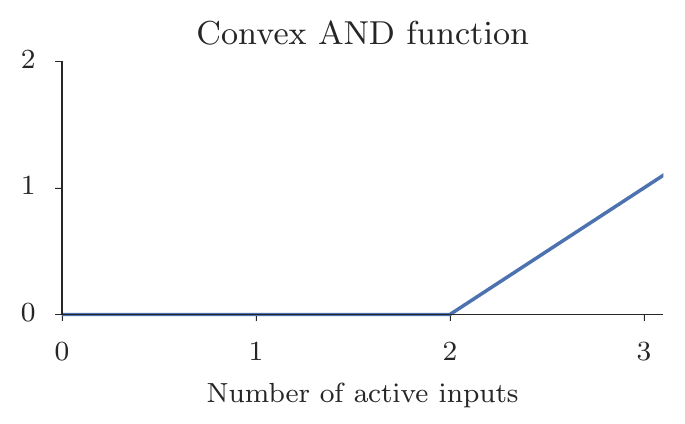}
\caption{Function $\phi(\mathbf{x}^T \mathbf{w} + b), w_i=1, b=-2$}
\label{and}
\end{subfigure}
\begin{subfigure}[b]{0.49\textwidth}
	\includegraphics[width=\linewidth]{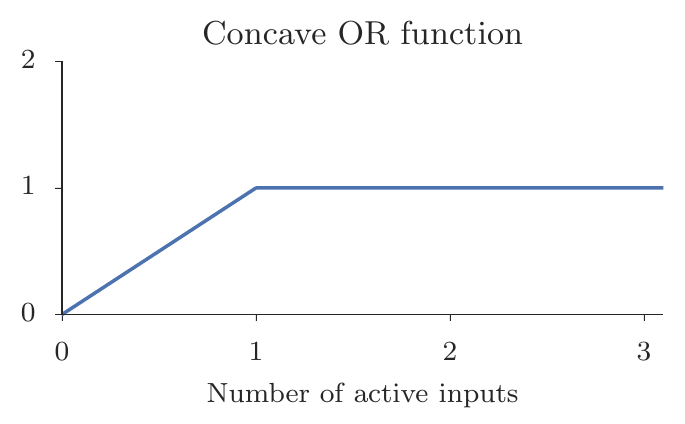}
\caption{Function $1-\phi(\mathbf{x}^T \mathbf{w} + b), w_i=-1, b=1$}
\label{or}
\end{subfigure}
\caption{Conceptual illustration of linear rectifier unit, $\phi(\cdot)$, performing 
	logical AND and OR
	operations for three-element binary input $\mathbf{x}, x_i \in \{0,1\}$
	after the affine transform
	$z = \mathbf{x}^T \mathbf{w} + b = x_0 w_0 + x_1 w_1 + x_2 w_2 + b$.
	AND operations has positive weights and convex form,
	whereas OR operation has negative weights and concave functional form.
	}
\label{andor}
\end{center}
\end{figure}

A striking feature of these plots is that the most invariant features
tend to have all negative weights. Since the nonlinearity $\phi(x)$ on
each layer was the rectified linear unit, a convex function which
saturates on the negative side, negative tied weights mean that the
network flipped these functions into $-\phi(-x)$, that is, concave
functions that saturate on the positive side. It therefore looks like
the network learned to add a layer of concave OR-like
invariance-generating poolings on top of convex AND-like coincidence
detectors. Forming convex AND using positive weights and concave OR
using negative weights is demonstrated respectively as truth tables in
Table~\ref{and_table} and Table~\ref{or_table}, and the geometric
form is illustrated in Figure~\ref{and} and Figure~\ref{or}.

\subsection{Invariance}
\label{sec:other_invariances}
\begin{figure}[tb]
\begin{center}

\begin{subfigure}[b]{\textwidth}
	\includegraphics[width=0.3\textwidth]{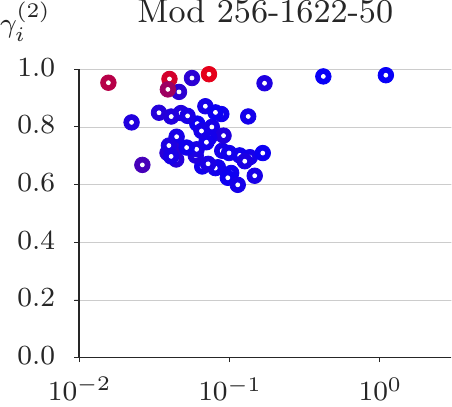}
	\includegraphics[width=0.3\textwidth]{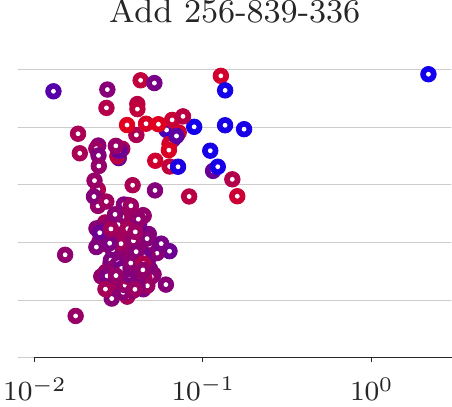}
	\includegraphics[width=0.3\textwidth]{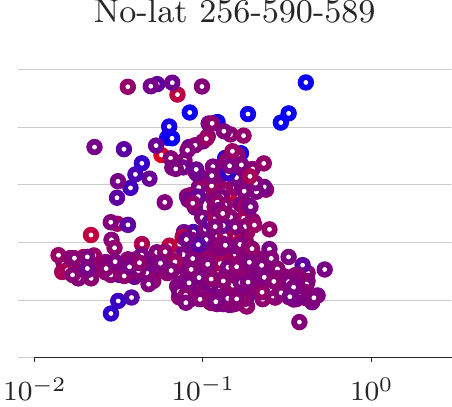}
    \caption{Rotation invariance}
    \label{invariance_rotation}
\end{subfigure}

\vspace{1cm}

\begin{subfigure}[b]{\textwidth}
	\includegraphics[width=0.3\textwidth]{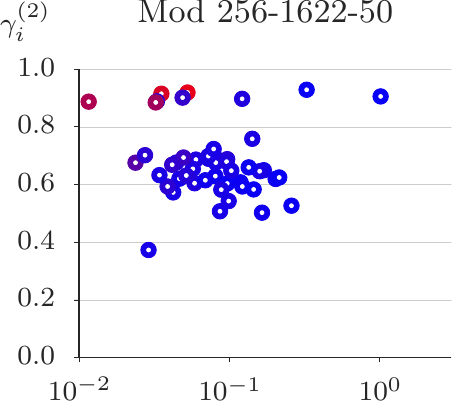}
	\includegraphics[width=0.3\textwidth]{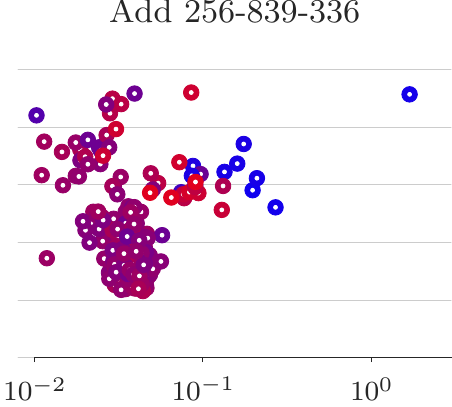}
	\includegraphics[width=0.3\textwidth]{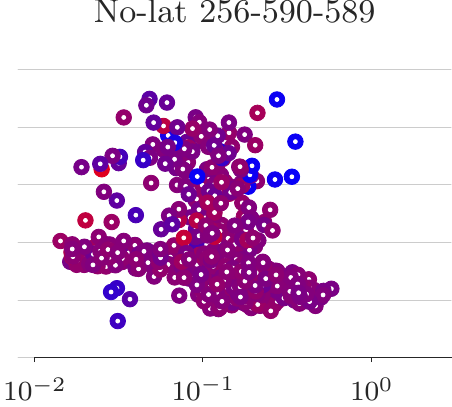}
    \caption{Scaling invariance}
    \label{invariance_scale}
\end{subfigure}

\vspace{1cm}

\begin{subfigure}[b]{\textwidth}
	\includegraphics[width=0.3\textwidth]{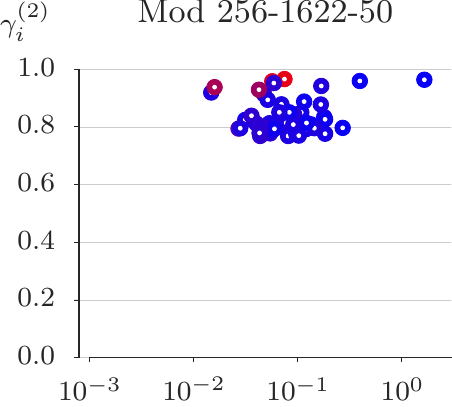}
	\includegraphics[width=0.3\textwidth]{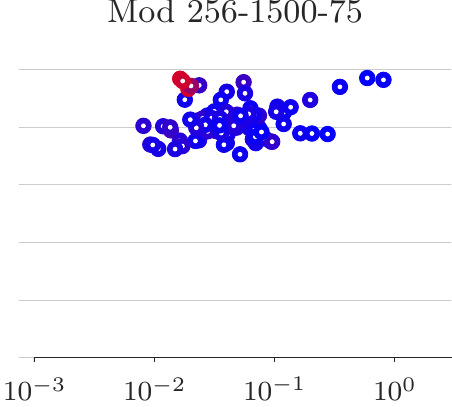}
	\includegraphics[width=0.3\textwidth]{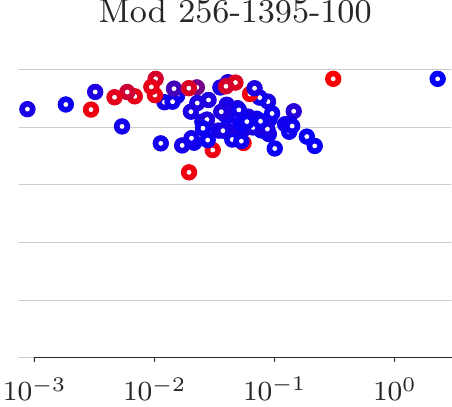}
    \caption{Translation invariance for Mod models with larger $\alpha$}
    \label{invariance_size}
\end{subfigure}
\end{center}
\caption{Various invariance measures for neurons of $h^{(2)}$ as a function of
significance for CIFAR-10.
Color indicates the average sign of the connected weights from negative
(blue) to positive (red).
Best viewed in color. See Section~\ref{sec:invariance} for more details.}
\end{figure}

We also studied invariance of the second layer neurons for scaling
and rotation transformations using the same invariance measure as
defined in Section~\ref{sec:invariance} for translation. We formed
sets of 16 samples by scaling CIFAR-10 images with a zoom
factors $0.6, 0.65, \dots, 1.35$ and rotating images $-32 \degree,
-28 \degree, \dots, 28 \degree$ for scaling and
rotation invariance experiments, respectively. The results are shown in
Figure~\ref{invariance_rotation} and Figure~\ref{invariance_scale}
and are similar to the translation invariance in Figure~\ref{invariance}.

Figure \ref{invariance_size} illustrates the impact of increasing
$\alpha$, i.e. increasing the size of the second layer.
Notably all second layer neurons stay highly invariant even
when the layer size is increased.

\begin{table}[p]
\caption{Truth table for logical AND operation given three-element binary
input vector $\mathbf{x}$ and a single perceptron with ReLU activation function $\phi(\cdot)$.
}
\begin{center}
$\phi(z)=\phi(\mathbf{x}^T \mathbf{w} + b) =
\phi(x_0 w_0 + x_1 w_1 + x_2 w_2 + b)$
\vspace{0.2cm}

\begin{tabular}{ccc|c|c||c|c||c}
 $x_0$ & $x_1$ & $x_2$ & $w_i$ & b & z & $\phi(z)$ & AND \\ \hline
 0 & 0 & 0 & 1 & -2 & -2 & 0 & 0 \\
 0 & 0 & 1 & 1 & -2 & -1 & 0 & 0 \\
 0 & 1 & 0 & 1 & -2 & -1 & 0 & 0 \\
 0 & 1 & 1 & 1 & -2 &  0 & 0 & 0 \\
 1 & 0 & 0 & 1 & -2 & -1 & 0 & 0 \\
 1 & 0 & 1 & 1 & -2 &  0 & 0 & 0 \\
 1 & 1 & 0 & 1 & -2 &  0 & 0 & 0 \\
 1 & 1 & 1 & 1 & -2 &  1 & 1 & 1 \\
\end{tabular}
\end{center}
\label{and_table}
\end{table}

\begin{table}[p]
\caption{Truth table for logical OR operation given three-element binary
input vector $\mathbf{x}$ and a single perceptron with ReLU activation function $\phi(\cdot)$.
}
\begin{center}
$\phi(z)=\phi(\mathbf{x}^T \mathbf{w} + b) =
\phi(x_0 w_0 + x_1 w_1 + x_2 w_2 + b)$
\vspace{0.2cm}

\begin{tabular}{ccc|c|c||c|c|c||c|c}
 $x_0$ & $x_1$ & $x_2$ & $w_i$ & b & z & $\phi(z)$ & $1-\phi(z)$ & NOR & OR \\ \hline
 0 & 0 & 0 & -1 & 1 &  1 & 1 & 0 & 1 & 0 \\
 0 & 0 & 1 & -1 & 1 &  0 & 0 & 1 & 0 & 1 \\
 0 & 1 & 0 & -1 & 1 &  0 & 0 & 1 & 0 & 1 \\
 0 & 1 & 1 & -1 & 1 & -1 & 0 & 1 & 0 & 1 \\
 1 & 0 & 0 & -1 & 1 &  0 & 0 & 1 & 0 & 1 \\
 1 & 0 & 1 & -1 & 1 & -1 & 0 & 1 & 0 & 1 \\
 1 & 1 & 0 & -1 & 1 & -1 & 0 & 1 & 0 & 1 \\
 1 & 1 & 1 & -1 & 1 & -2 & 0 & 1 & 0 & 1 \\
\end{tabular}
\end{center}
\label{or_table}
\end{table}

\begin{figure}[tb] \begin{center}
       \includegraphics[width=0.7\linewidth]{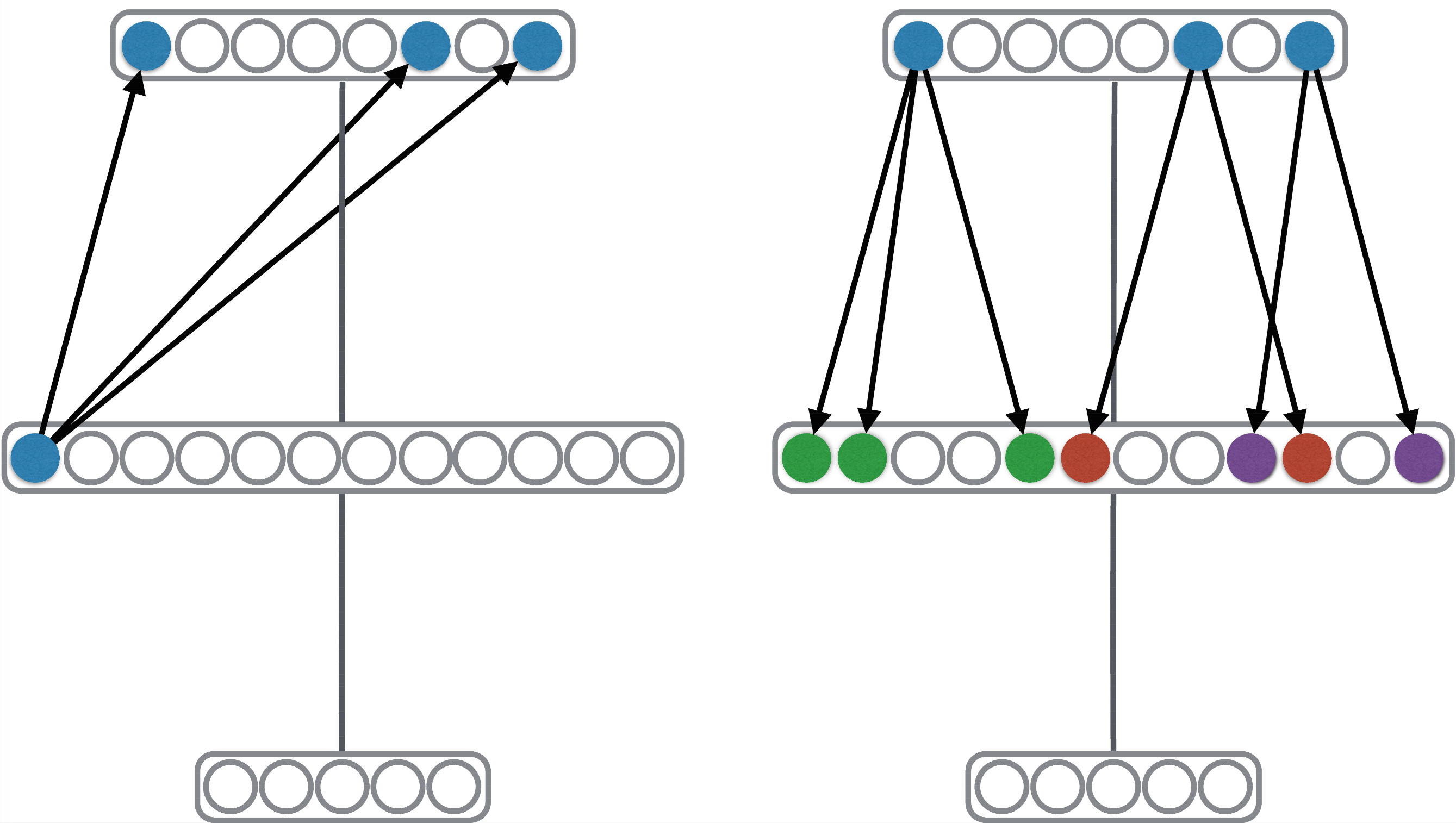}
\label{feature_lookup}
\end{center}
\caption{Method for selecting pooling groups including a given $\h^{(1)}$
    neuron contains two phases.  First, follow the strongest links up to the
    second layer to identify 3 pooling groups the first-layer neuron belongs to
    (left).
    Second, visualize each pooling group by identifying the Layer 1 neurons
    that have the strongest links back from each Layer 2 neuron (right). In
    this example, the first pooling group contains neurons marked with green,
    the second pooling group with red, and the third consist of purple neurons.
    Colors correspond to rows in Figure~\ref{filters} and both phases were
    performed for each set (a)-(d).
}
\label{pooling_explained}
\end{figure}

\end{document}

%% file: results.tex
Dataset & Model & Layer sizes & Ratio $\alpha$ & Min cost $\pm$ std & Invariance, $\gamma^{(l)}$ & \\ \hline
CIFAR-10 & Linear  & 256 & & 0.20316 $\pm$ 0.00011 & NA & \\
CIFAR-10 & No-lat  & 256-1948 & 0.00 & 0.16144 $\pm$ 0.00007 & 0.20, 0.29 & \\
CIFAR-10 & Add/Mod & 256-1937 & 0.00 & 0.15793 $\pm$ 0.00007 & 0.20, 0.29 & \\
CIFAR-10 & No-lat  & 256-590-589 & 1.00 & 0.17595 $\pm$ 0.00034 & 0.20, 0.23, 0.30 & \\
CIFAR-10 & Add  & 256-839-336 & 0.40 & 0.15734 $\pm$ 0.00015 & 0.20, 0.27, 0.33 & \\
CIFAR-10 & Mod  & 256-1622-50 & 0.03 & \textbf{0.15086} $\pm$ 0.00022 & 0.20, 0.31, 0.84 & \\
\hline
O\&F & Linear  & 256 & & 0.14686 $\pm$ 0.00010 & NA & \\
O\&F & No-lat  & 256-1948 & 0.00 & 0.12290 $\pm$ 0.00008 & 0.16, 0.24 & \\
O\&F & Add/Mod & 256-1937 & 0.00 & 0.11891 $\pm$ 0.00006 & 0.16, 0.25 & \\
O\&F & No-lat  & 256-512-718 & 1.40 & 0.12446 $\pm$ 0.00015 & 0.16, 0.19, 0.27 & \\
O\&F & Add  & 256-1061-212 & 0.20 & 0.11723 $\pm$ 0.00009 & 0.16, 0.25, 0.41 & \\
O\&F & Mod  & 256-1395-100 & 0.07 & \textbf{0.11234} $\pm$ 0.00005 & 0.16, 0.24, 0.95 & \\